\documentclass[journal,twoside,web]{ieeecolor}
\usepackage[OT1]{fontenc}
\usepackage{generic}
\usepackage{cite}
\usepackage{amsmath,amssymb,amsfonts}
\usepackage{algorithmic}
\usepackage{graphicx}
\usepackage{algorithm,algorithmic}
\usepackage{hyperref}
\hypersetup{hidelinks=true}
\usepackage{textcomp}
\usepackage{booktabs}
\usepackage{float}
\usepackage{subcaption} % preamble
\def\BibTeX{{\rm B\kern-.05em{\sc i\kern-.025em b}\kern-.08em
    T\kern-.1667em\lower.7ex\hbox{E}\kern-.125emX}}
\begin{document}
\title{MURFormer: Attention-Based Lightweight Multi-Scale Uncertainty-Guided Refinement Framework for PPG Denoising}
\author{Haram Kwon, Chaeeun Hwang, Seayul Choi, and Sungho Jo, \IEEEmembership{Senior Member, IEEE}
\thanks{}
\thanks{This work involved human subjects or animals in its research. Approval of all ethical and experimental procedures and protocols was granted by Institutional Review Board, Korea Advanced Institute of Science and Technology (KAIST) under Application No. KAISTIRB-2025-279, and performed in line with the Declaration of Helsinki.}
\thanks{Haram Kwon, Chaeeun Hwang, Seayul Choi, and Sungho Jo are with the School of Computing, Korea Advanced Institute of Science and Technology, Daejeon 34141, South Korea (e-mail: hyrumk@kaist.ac.kr; navy3690@kaist.ac.kr; sycseayul@kaist.ac.kr; shjo@kaist.ac.kr)}}

\maketitle

\begin{abstract}
% TODO: github repo 공개하기
Photoplethysmography (PPG) is a widely used physiological signal for its convenience but is highly susceptible to motion artifacts that degrade waveform quality and limit downstream analysis. This paper proposes MURFormer, a lightweight multi-scale uncertainty-guided refinement framework for PPG denoising through noisy-to-clean PPG mapping. MURFormer employs a newly proposed Multi-Scale Attention-Conv Fusion (MACF) block, which is designed to efficiently capture multi-scale temporal dynamics of PPG, as the fundamental building block and performs uncertainty-guided two-pass refinement prioritizing unreliable regions. The method was evaluated on the publicly available CIME dataset and on a Daily Activities dataset collected in this work, comprising paired noisy-clean PPG recordings from 24 subjects. Experimental results show that MURFormer improves both signal quality and physiological fidelity over the raw noisy input, yielding 129\% and 105\% relative signal-to-noise ratio (SNR) improvements, and 66\% and 81\% relative reductions in heart-rate mean absolute error (MAE), on the Daily Activities and CIME datasets, respectively. MURFormer also outperforms an existing denoising baseline on both datasets with a relatively small parameter count while preserving beat-level timing and waveform morphology. In addition, the model maintains strong denoising performance across varying input window lengths, suggesting that MURFormer is an effective and robust framework for practical PPG denoising. The implementation of MURFormer is publicly available at \url{https://github.com/k-hram/MURFormer.git}

\end{abstract}


\section{Introduction}
\label{sec:introduction}

\IEEEPARstart{P}{hotoplethysmography} (PPG) is a simple, cost-effective, and noninvasive optical sensing technique commonly used to detect changes in blood volume in the microvascular bed of tissue \cite{allen2007photoplethysmography, zhang2014troika}. PPG has been extensively adopted in various physiological monitoring tasks such as heart rate estimation, pulse interval analysis\cite{charlton20232023}, and vascular assessments through wearable or clinical monitoring systems, due to its convenient acquisition \cite{castaneda2018review, charlton2022wearable, ferizoli2024arterial}. However, its high vulnerability to motion artifacts often limits the practical utility of PPG, particularly in daily-life settings, where signal corruption can distort waveform morphology and degrade the reliability of PPG-based downstream analysis \cite{he2019robust, maity2022ppgmotion}.

This vulnerability is particularly problematic in wearable scenarios, where body movement, sensor displacement, and varying contact conditions frequently contaminate the measured signal. Under such conditions, motion artifacts not only affect visual signal quality but can also alter peak timing, local morphology, and pulse-to-pulse consistency, which complicate physiological analysis in PPG.

A large body of prior work has therefore addressed PPG denoising and motion-artifact reduction. Traditional methods often relied on signal processing techniques such as wavelet transform\cite{joseph2014photoplethysmogram,ahmed2023deep}, adaptive filtering\cite{ram2011novel,raghuram2016use,tanweer2017motion}, and decomposition\cite{ye2016combining,ye2017robust,gupta2023exploiting}. More recent work has explored learning-based approaches such as autoencoder-based models\cite{mohagheghian2023noise, jain2023self}, generative adversarial networks (GANs)\cite{wang2022ppg,zheng2022ppg,afandizadeh2023accurate, sawangjai2023ppganet, sawangjai2025removal}, mamba\cite{chiu2025robust}, transformer\cite{kwon2022preeminently}, learned convolutional sparse coding\cite{basso2026reduction}, and diffusion\cite{xia2025effective,liu2026flexima}. These studies have demonstrated that learning-based approaches can effectively reduce noise by reconstructing a clean PPG signal from a corrupted input. However, the way in which noisy-clean PPG pairs are constructed remains an important limitation. Several studies have trained denoising models by adding extracted or simulated noise to clean PPG signals\cite{mohagheghian2023noise,chiu2025robust,basso2026reduction}, using self-supervised erasing of clean segments\cite{jain2023self}, or adopting unpaired noisy-to-clean translation\cite{afandizadeh2023accurate}. Although these strategies reduce the need for paired recordings, they may not faithfully reproduce real motion artifacts arising from sensor displacement, skin deformation, and motion-related physiological changes, potentially limiting generalization to wearable settings.

To address this limitation, recent work by Xia et al.\cite{xia2025effective} and Liu et al.\cite{liu2026flexima} employ noisy-clean PPG pairs for supervised reconstruction. These studies have demonstrated that paired-data approaches can effectively reduce noise by reconstructing a clean PPG waveform from a corrupted input. Nevertheless, existing paired-data denoisers often rely on relatively complex generative or multi-module pipelines. Therefore, this work aims to retain the benefits of paired noisy-clean supervision while introducing a lightweight and parameter-efficient denoising framework.

% Propose model
%  MURFormer intro v3
This study proposes MURFormer (Multi-Scale Uncertainty-Guided Refinement Former), a lightweight uncertainty-guided two-pass refinement framework trained on paired noisy-clean PPG recordings. Although several prior methods incorporate motion reference signals such as accelerometer or gyroscope measurements \cite{wu2017implementation,lee2018motion,kasnesis2023feature}, MURFormer uses only PPG as input and does not require additional motion signals to reconstruct clean signals. While excluding IMU signals may remove useful motion-reference information, this design makes the framework applicable to PPG-only recordings and avoids additional synchronization requirements. The model combines the newly proposed Multi-Scale Attention-Convolution Fusion (MACF) block with a three-stage training strategy for reconstruction, uncertainty estimation, and refinement. Unlike prior artifact-detection-and-reconstruction frameworks that use detection mainly for artifact removal preprocessing\cite{afandizadeh2023accurate,jain2023self,liu2026flexima}, MURFormer uses uncertainty to further refine the signal by prioritizing regions that remain unreliable after initial reconstruction.

% About the evaluation
Experimental results on both datasets show that MURFormer substantially improves signal quality and physiological fidelity over noisy inputs, outperforms a recent diffusion-based baseline\cite{xia2025effective} with fewer parameters, and maintains robust performance across varying test-time input lengths.

% Just an image of MURFormer and an MACF block %
\begin{figure*}[t]
    \centering
    \includegraphics[width=\textwidth]{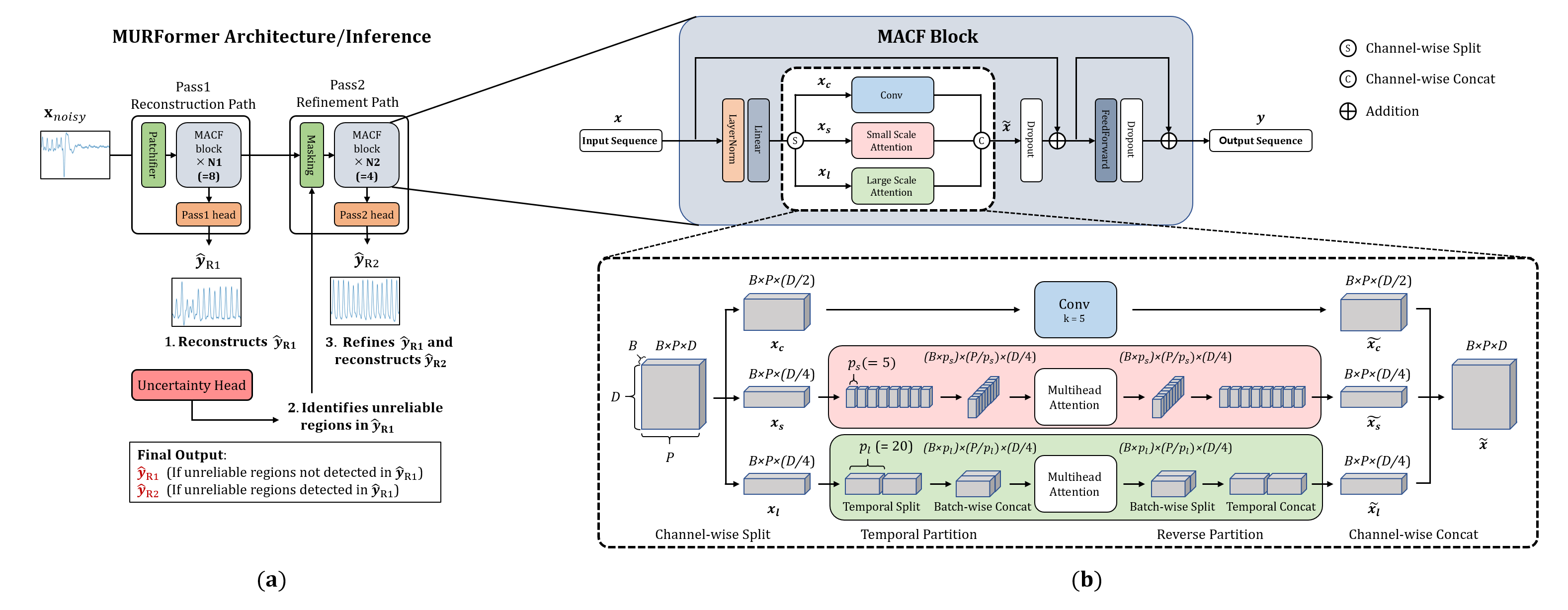}
    \caption{(a) MURFormer architecture and inference workflow. (b) MACF block architecture. The top panel presents the overall architecture of an MACF block, and the bottom panel illustrates the multi-scale attention mechanism in its triple-branch structure.}
    \label{fig:murformer_macf}
\end{figure*}

In summary, the main contributions of this work are as follows:
\begin{enumerate}
    \item We propose MURFormer, a lightweight PPG denoising framework that achieves strong denoising performance.
    \item We introduce the MACF block, a parameter-efficient encoder designed to capture local and multi-scale temporal characteristics of PPG.
    \item We present an uncertainty-guided two-pass refinement strategy that improves reconstruction quality.
\end{enumerate}

\section{Methods}
\label{sec:methods}

\subsection{Model Overview}
\label{subsec:model_overview}
% version2 (shorter, hopefully)
MURFormer is designed to denoise motion-corrupted PPG signals using paired noisy-clean recordings. Given a noisy PPG segment $\mathbf{x}_{noisy} \in \mathbb{R}^{T}$, the model reconstructs its clean counterpart $\mathbf{y}_{clean} \in \mathbb{R}^{T}$ through a two-pass framework composed of first-pass reconstruction, patch-level uncertainty estimation, and second-pass refinement.

\begin{figure*}[t]
    \centering
    \includegraphics[width=\textwidth]{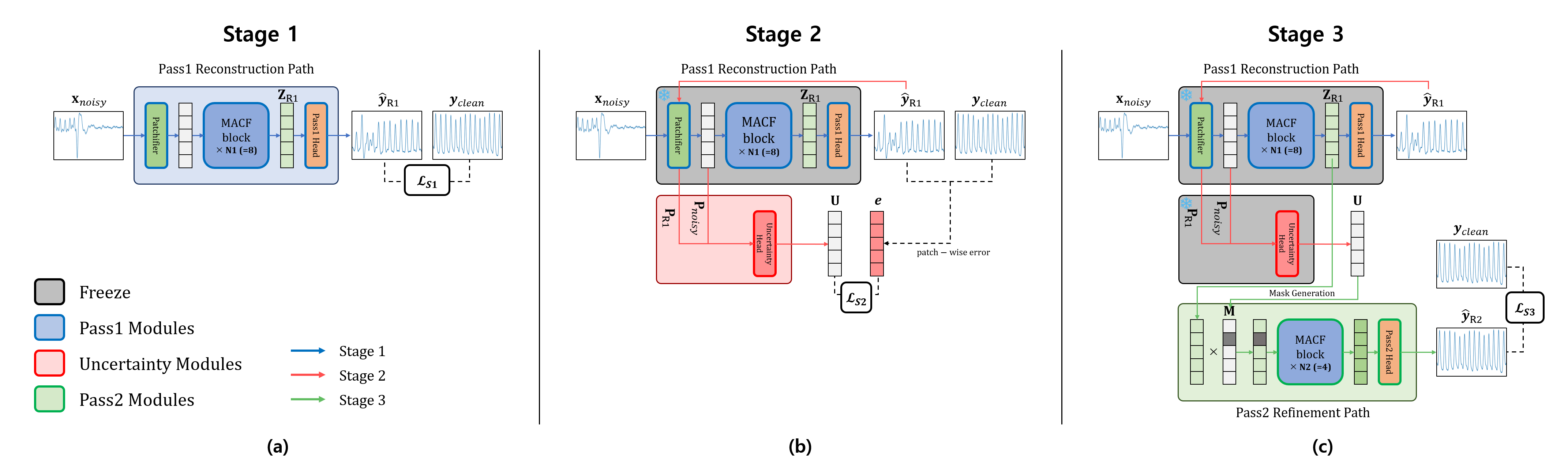}
    \caption{Three-stage training strategy and the model overview. Frozen layers at each stage are overlaid with gray squares. (a) Stage 1 overview, only the parameters of the pass1 reconstruction path are updated. (b) Stage 2, where only the parameters of the uncertainty head are updated. (c) Stage 3, only the parameters of the pass2 refinement path are updated.}
    \label{fig:training}
\end{figure*}

The architecture of MURFormer is shown in Fig.~\ref{fig:training}c and Fig.~\ref{fig:murformer_macf}a. The pass1 reconstruction path, consisting of a patchifier, MACF blocks, and a pass1 reconstruction head, converts the noisy input PPG into an initial denoised signal. The uncertainty head predicts a patch-level uncertainty map that indicates which temporal regions of the first-pass output are likely to remain unreliable. The pass2 refinement path, consisting of MACF blocks and a pass2 reconstruction head, then takes a masked version of the pass1 latent representation and prioritizes the refinement of unreliable patches to produce the final denoised output. The remainder of this section describes the MACF block, the three-stage training procedure, and the two-pass inference workflow in detail.

\subsection{MACF Block}
\label{subsec:macf_block}

Fig.~\ref{fig:murformer_macf}b illustrates the proposed Multi-Scale Attention-Convolution Fusion (MACF) block, which serves as the basic encoder block of MURFormer. The residual-style structure of an MACF block resembles that of a standard Transformer encoder block\cite{vaswani2017attention}, but differs in that the input is processed through a triple-branch structure, intended to capture both local waveform morphology and temporal dependencies at multiple scales. Each branch is assigned to a different operation: local convolution, small-scale attention, and large-scale attention. This design enables the block to model complementary temporal patterns while using fewer parameters than applying attention to the full embedding dimension.

Let the input sequence be $\mathbf{x}\in\mathbb{R}^{B\times P\times D}$, where $B$, $P$, and $D$ denote the batch size, the number of patch tokens, and the embedding dimension, respectively. The overall flow of the MACF block can be written as

\begin{align}
[\mathbf{x}_c,\mathbf{x}_s,\mathbf{x}_l] &= \mathrm{Split}(\mathrm{Linear}(\mathrm{LN}(\mathbf{x}))), \\
\tilde{\mathbf{x}} &= \mathrm{Concat}(\mathrm{Conv}(\mathbf{x}_c),\mathrm{Attn}_s(\mathbf{x}_s),\mathrm{Attn}_l(\mathbf{x}_l)), \\
\mathbf{z} &= \mathbf{x} + \mathrm{Dropout}(\tilde{\mathbf{x}}), \\
\mathbf{y} &= \mathbf{z} + \mathrm{Dropout}(\mathrm{FFN}(\mathbf{z})),
\end{align}
where $\mathrm{LN}(\cdot)$ denotes layer normalization, $\mathrm{Linear}(\cdot)$ denotes a linear projection, $\mathrm{Split}(\cdot)$ denotes channel-wise feature splitting, $\mathrm{Conv}(\cdot)$ denotes the convolution branch, $\mathrm{Attn}_s(\cdot)$ and $\mathrm{Attn}_l(\cdot)$ denote small-scale and large-scale attention branches, respectively, $\mathrm{Concat}(\cdot)$ denotes channel-wise concatenation, and $\mathrm{FFN}(\cdot)$ denotes the position-wise feed-forward network with an expansion ratio of 2.

As shown in Fig.~\ref{fig:murformer_macf}b, an MACF block input sequence $\mathbf{x}\in\mathbb{R}^{B\times P\times D}$ is first linearly projected while preserving its shape, and the resulting sequence is then split along the channel dimension into
\begin{equation} 
\mathbf{x}_c \in \mathbb{R}^{B\times P\times (D/2)},
\mathbf{x}_s \in \mathbb{R}^{B\times P\times (D/4)},
\mathbf{x}_l \in \mathbb{R}^{B\times P\times (D/4)}. 
\end{equation}

The first branch is a convolution branch that takes $\mathbf{x}_c$ and applies 1-D convolution of kernel size 5 along the temporal axis while preserving the input shape through padding, producing $\tilde{\mathbf{x}_c}$.

The second branch is a small-scale attention branch, in which temporal partition, multi-head attention, and reverse partition are applied sequentially. For temporal partition, the input $\mathbf{x}_s$ is partitioned along the temporal dimension using partition size $p_s$ (Eq. \eqref{eq:temporal_split}), a small-scale partition size, followed by batch-wise concatenation (Eq. \eqref{eq:batch_concat}). Thus,
\begin{align}
\mathbf{x}_s \in \mathbb{R}^{B\times P\times (D/4)}
&\rightarrow
\mathbb{R}^{B\times (P/p_s)\times p_s\times (D/4)} \label{eq:temporal_split} \\
&\rightarrow
\mathbb{R}^{(B\times p_s)\times (P/p_s)\times (D/4)}. \label{eq:batch_concat}
\end{align}
Multi-head attention is then applied to the partitioned sequence, followed by a reverse partition, which restores the original sequence layout and produces $\tilde{\mathbf{x}_s}$.
\begin{align}
\mathbb{R}^{(B\times p_s)\times (P/p_s)\times (D/4)}
&\rightarrow
\mathbb{R}^{B\times (P/p_s)\times p_s\times (D/4)} \label{eq:temporal_concat} \\
&\rightarrow
\tilde{\mathbf{x}_s} \in \mathbb{R}^{B\times P\times (D/4)}.
\end{align}
The third branch is a large-scale attention branch that follows the same procedure as the second branch, but uses a different partition size, $p_l$, and a separate multi-head attention module of the same structure, producing $\tilde{\mathbf{x}_l}$.

% [시율] 정말 사소하지만, partition size가 그림에는 p_S, p_L(대문자)로 나와있지만 본문에는 p_s, p_l (소문자)로 쓰여있습니다..! -> 감사합니다! 확인하고 고쳤어요!

The outputs of three branches $\tilde{\mathbf{x}_c}$, $\tilde{\mathbf{x}_s}$, and $\tilde{\mathbf{x}_l}$ are concatenated along the channel dimension to recover the original embedding size, yielding a fused representation $\tilde{\mathbf{x}}\in\mathbb{R}^{B\times P\times D}$. Dropout is then applied followed by residual addition with the block input. The resulting feature is further processed by a position-wise feed-forward network, with another dropout and residual addition following to produce the output $\mathbf{y}$.

Since attention is applied only to partial channel subspaces, the MACF block uses fewer attention parameters than a standard Transformer encoder with the same input shape. With the adopted patch configuration, each token corresponds to 0.25\,s of the input signal. The value of $p_s$ is set to 5, aggregating information over 1.25\,s of the input. This temporal span is slightly longer than a single pulse interval at 60 bpm, allowing the small-scale attention branch to apply multi-head attention over approximately one to two cardiac cycles. We further set the value of $p_l$ to 20, so that the large-scale attention branch covers 5\,s, and captures dependencies across multiple pulses.

\subsection{Three-Stage Training Strategy} 
\label{subsec:three_stage_training}

A three-stage training strategy is designed to train MURFormer to first learn a reliable reconstruction path, then to estimate a patch-wise reconstruction uncertainty, and to guide refinement toward unreliable regions. At each stage, only modules that are relevant to the current objective remain unfrozen in order to allow the model to learn each function progressively while preserving previously learned components. An overview of the training strategy is shown in Fig.~\ref{fig:training}.

\subsubsection*{Stage 1: Direct Noisy-to-Clean Reconstruction}
In Stage 1, the noisy input $\mathbf{x}_{noisy}$ is converted into patch tokens $\mathbf{P}_{noisy}$. Then it is processed by the MACF backbone and the pass1 reconstruction head, producing the first-pass denoised signal. The objective of this stage is to minimize the reconstruction loss $\mathcal{L}_{\text{S1}}$.

The Stage 1 reconstruction loss is defined as
\begin{equation}
\label{eq:stage1_loss}
\mathcal{L}_{\text{S1}}
=
\mathcal{L}_{\text{char}}
+\alpha\,\mathcal{L}_{\text{spec}}
+\beta\,\mathcal{L}_{\nabla}
+\gamma\,\mathcal{L}_{\nabla^2},
\end{equation}
where $\mathcal{L}_{\text{char}}$ is the Charbonnier loss, $\mathcal{L}_{\text{spec}}$ is the spectral loss that penalizes mismatches in the frequency domain, and $\mathcal{L}_{\nabla}$ and $\mathcal{L}_{\nabla^2}$ enforce consistency in the first and second-order temporal derivatives $(\alpha,\beta,\gamma \in \mathbb{R})$. These loss functions are expressed as
% ---------------------
% Stage 1 Loss (Combined Reconstruction)
% ---------------------
\begin{align}
\mathcal{L}_{\text{char}}
&=
\frac{1}{BT}
\sum_{b,t}
\sqrt{\left((\hat{\mathbf{y}}_{R1})_{b,t}-(\mathbf{y}_{clean})_{b,t}\right)^2+\epsilon^2},
\\
\mathcal{L}_{\text{spec}}
&=
\frac{
\left\|
\left|\mathrm{rFFT}(\hat{\mathbf{y}}_{R1})\right|
-
\left|\mathrm{rFFT}(\mathbf{y}_{clean})\right|
\right\|_2
}{
N_{\mathrm{fft}}
},
\\
\mathcal{L}_{\nabla}
&=
\frac{1}{BT}
\sum_{b,t}
\sqrt{\left(\nabla (\hat{\mathbf{y}}_{R1})_{b,t}-\nabla (\mathbf{y}_{clean})_{b,t}\right)^2+\epsilon^2},
\\
\mathcal{L}_{\nabla^2}
&=
\frac{1}{BT}
\sum_{b,t}
\sqrt{\left(\nabla^2 (\hat{\mathbf{y}}_{R1})_{b,t}-\nabla^2 (\mathbf{y}_{clean})_{b,t}\right)^2+\epsilon^2}.
\end{align}

Here, $B$ and $T$ denote the batch size and sequence length, respectively. $\hat{\mathbf{y}}_{R1}$ and $\mathbf{y}_{clean}$ are the reconstructed and target clean reference signal, and $\epsilon$ is a small constant in the Charbonnier term. In $\mathcal{L}_{\text{spec}}$, real-valued fast Fourier transform (rFFT) is applied along the time axis of each sample. It is then normalized by $N_{\mathrm{fft}} = \Omega B$, where $\Omega$ denotes the number of non-redundant frequency bins produced by the rFFT.

Only the parameters of the patchifier, the pass1 MACF backbone, and the pass1 reconstruction head are updated in this stage.

\subsubsection*{Stage 2: Patch-Level Uncertainty Estimation}
\label{subsubsec:stage2}
In Stage 2, the pass1 reconstruction path trained during Stage 1 is frozen, and only the uncertainty head is trained. For uncertainty estimation, $\hat{\mathbf{y}}_{R1}$ is first passed through the Stage 1 patchifier to obtain its patch-level representation $\mathbf{P}_{R1}$. $\mathbf{P}_{R1}$ and $\mathbf{P}_{\mathrm{noisy}}$ from Stage 1 are concatenated along the feature dimension to form the input to the uncertainty head. The uncertainty head consists of two linear layers with LayerNorm and ReLU activation between them, followed by a Softplus activation.

Stage 2 uses Gaussian negative log-likelihood loss to train the model to estimate patch-level aleatoric uncertainty\cite{kendall2017uncertainties}. Stage 2 loss can be expressed as 

% ---------------------
% Stage 2 Loss
% ---------------------
\begin{align}
e_{b,p}
&=
\frac{1}{s}
\sum_{t \in \mathcal{P}_p}
\left((\hat{\mathbf{y}}_{R1})_{b,t}-(\mathbf{y}_{clean})_{b,t}\right)^2,
\\
\mathcal{L}_{\text{S2}}
&=
\frac{1}{BP}
\sum_{b,p}
\left(
\frac{e_{b,p}}{2U_{b,p}}
+\frac{1}{2}\log U_{b,p}
\right)
\label{eq:stage2_loss}.
\end{align}

where $B$, $P$, and $s$ denote the batch size, the number of temporal patches per sample, and the number of time points in each patch, respectively. $(\hat{\mathbf{y}}_{R1})_{b,t}$ and $(\mathbf{y}_{\mathrm{clean}})_{b,t}$ denote the values at time point $t$ of the $b$-th first-pass denoised output and the corresponding clean reference signal, respectively. $e_{b,p}$ is the patch-level MSE for patch $p$, and $U_{b,p}$ denotes the $(b,p)$-th element of the predicted uncertainty map $\mathbf{U}$, with $U_{b,p}>0$.

\subsubsection*{Stage 3: Uncertainty-Guided Refinement Learning}
In Stage 3, the pass1 reconstruction path and the uncertainty head are frozen, and only the parameters of the pass2 refinement path are updated. Stage 3 leverages the outputs of both Stage 1 and Stage 2. It first takes the uncertainty map $\mathbf{U} \in \mathbb{R}^{B\times P}$ produced by the uncertainty head. Based on $\mathbf{U}$, a core binary mask $\textbf{I}$ is obtained by thresholding the uncertainty values with the threshold $\tau$ as
\begin{equation}
\label{eq:core_mask}
I_{b,p} = \mathbf{1}\!\left(U_{b,p} > \tau\right),
\qquad
\textbf{I} \in \{0,1\}^{B \times P}.
\end{equation}
Here, $b$ and $p$ denote batch and patch indices, respectively. Patches whose uncertainty exceeds the threshold are treated as unreliable. 

Masking is applied to $\mathbf{Z}_{R1}$, the first-pass latent representation, in two ways. The first approach is neighbor-expanded uncertainty masking, which serves as the primary masking policy used consistently during both training and inference. In this masking approach, core mask $I_{b,p}$ is expanded to 1-hop temporal neighbors as shown in Eq. \eqref{eq:neighbor_expand}.

% 1-hop temporal neighbor expansion
\begin{equation}
\label{eq:neighbor_expand}
M_{b,p} = I_{b,p-1} \,\lor\, I_{b,p} \,\lor\, I_{b,p+1},
\qquad
\mathbf{M} \in \{0,1\}^{B \times P}.
\end{equation}
The purpose of this masking expansion is to prevent a single patch masking from creating sharp transitions at patch boundaries. The second masking approach is random block masking, which is used only during training. For each sample, a target mask ratio is first sampled, and contiguous patch blocks with random start positions and lengths are then masked until the target masked-patch budget is reached. This strategy mainly serves as a fallback when the uncertainty-driven mask is empty.

Once $\mathbf{Z}_{R1}$ is masked, it is processed by the pass2 MACF backbone and the pass2 reconstruction head to produce the second-pass denoised signal. The objective of the Stage 3 refinement path is to first mask unreliable patches and refine such regions. Thus, the Stage 3 loss is defined as

% ---------------------
% Stage 3 Loss
% ---------------------
\begin{align}
\mathcal{L}_{\text{masked-MSE}}
&=
\frac{
\sum_{b,p}\sum_{t \in \mathcal{P}_p}
M_{b,p}\left((\hat{\mathbf{y}}_{R2})_{b,t}-(\mathbf{y}_{clean})_{b,t}\right)^2
}{
s \sum_{b,p} M_{b,p}
},
\\
\mathcal{L}_{\text{global-MSE}}
&=
\frac{1}{BT}
\sum_{b,t}
\left((\hat{\mathbf{y}}_{R2})_{b,t}-(\mathbf{y}_{clean})_{b,t}\right)^2,
\\
\mathcal{L}_{\text{S3}}
&=
\alpha_m\,\mathcal{L}_{\text{masked-MSE}}
+\mathcal{L}_{\text{global-MSE}}
\label{eq:stage3_loss}.
\end{align}
where $\mathcal{P}_p$ denotes the timepoints that belong to $p$-th patch, $\textbf{M}$ the binary mask, and $\hat{\mathbf{y}}_{R2}$ the pass2 denoised output.

$\mathcal{L}_{\text{masked-MSE}}$ concentrates supervision on patches selected for refinement, while $\mathcal{L}_{\text{global-MSE}}$ provides a global constraint on the full waveform to prevent discontinuities at the boundaries of masked and unmasked regions. $\mathcal{L}_{\text{masked-MSE}}$ and $\mathcal{L}_{\text{global-MSE}}$ are used together, with a larger weight assigned to $\mathcal{L}_{\text{masked-MSE}}$ ($\alpha_m$ = 5 in our experiments) to balance targeted refinement and overall signal fidelity.

\subsection{Two-Pass Inference}
\label{subsec:inference}

Fig.~\ref{fig:murformer_macf}a depicts the inference workflow of MURFormer. The noisy input is first processed by the pass1 reconstruction path to obtain the first-pass denoised output $\hat{\mathbf{y}}_{R1}$. A patch-level uncertainty map is then predicted and thresholded to identify unreliable regions as described in Section~\ref{subsubsec:stage2} and Eq. \eqref{eq:core_mask}. If no patch exceeds the uncertainty threshold $\tau$, the first-pass output $\hat{\mathbf{y}}_{R1}$ is used as the final denoised signal. Otherwise, the refinement mask is applied to the first-pass latent sequence $\mathbf{Z}_{R1}$. The masked latent sequence is subsequently processed by the pass2 refinement path and its output $\hat{\mathbf{y}}_{R2}$ is used as the final denoised signal.
% [시율] 정말정말 사소하긴 하지만... '...pass2 refinement path and its output is used..' -> '...pass2 refinement path and its output $\hat{\mathbf{y}}_{R2}$ is used..' 로 수정하는 것이 어떨지 조심스레 건의드립니다. --> 문장 구조상 그게 일관성 있어보여서 좋네요. 고쳤습니다!

\section{Experimental Setup}
\label{sec:expsetup}

\subsection{Dataset}
\label{subsec:dataset}
We used the publicly available CIME dataset and a Daily Activities dataset  collected in this study. These datasets include noisy-clean PPG pairs which are suitable for training MURFormer. Although the IEEE Signal Processing Cup 2015 Database (ISPC) contains noisy-clean PPG pairs, it was not used due to its limited size\cite{zhang2014troika}.

\subsubsection{CIME Dataset}
\label{subsubsec:cime}
For training and evaluation, we used the public dataset CIME\cite{xu2019photoplethysmography, xu2020stochastic}, which contains paired noisy and clean finger PPG recordings from 48 individuals. The signals were acquired simultaneously from the left and right index fingertips, with the left finger kept still and the right finger performing movements.

\subsubsection{Daily Activities Dataset}
\label{subsubsec:data_collection}

We collected a paired noisy-clean PPG dataset, referred to in this paper as the Daily Activities dataset, to support both supervised noisy-to-clean PPG denoising and evaluation under realistic daily-life motion conditions. Signals were recorded at 40 Hz using a custom watch-type device built with an Arduino ESP32-based TinyPICO board and the Gravity: Heart Rate Monitor Sensor for Arduino.

During recording, each subject wore the devices on both wrists. The dominant hand performed the assigned task while the non-dominant hand was kept still to provide a relatively clean reference signal. Thus, PPG obtained from the dominant hand was used as the noisy signal and the PPG obtained from the non-dominant hand was used as the clean signal.

PPG was collected from a total of 24 participants, all of whom provided informed consent prior to participating in the data collection. Each subject participated in two recording sessions conducted on different days. In each session, eight tasks were performed in the fixed order of \textit{rest}, \textit{writing}, \textit{typing}, \textit{gaming}, \textit{walking}, \textit{exercise}, \textit{eating}, and \textit{cleaning}. Each task lasted 5 minutes, and a 1 minute rest period was provided between tasks. Most tasks were performed in a seated position, whereas \textit{walking} and \textit{exercise} were performed while standing. The tasks were designed to induce motion artifacts representative of daily activities while keeping the non-dominant hand as still as possible. The activity tasks included writing, one-handed typing, touchscreen gaming, walking in place with dominant-arm swing, one-sided stretching exercise, eating with chopsticks, and cleaning-related object manipulation.

The study protocol was approved by the Institutional Review Board of Korea Advanced Institute of Science and Technology (KAIST) under Application No. KAISTIRB-2025-279 on November 25, 2025.

%The tasks were designed to induce diverse motion artifacts likely to occur in daily activities. Subjects were instructed to use only the dominant hand and to keep the non-dominant hand as still and relaxed as possible. Each task was performed as follows. In the \textit{rest} task, subjects remained seated with both forearms placed on the desk. In the \textit{writing} task, subjects wrote a given text using only the dominant hand. In the \textit{typing} task, subjects typed a given text using only the dominant hand. In the \textit{gaming} task, subjects played a touchscreen whack-a-mole game using the dominant hand. In the \textit{walking} task, subjects walked in place while swinging only the dominant arm and keeping the non-dominant arm fixed. In the \textit{exercise} task, subjects followed a stretching video using only the dominant side of the body while keeping the other side still. In the \textit{eating} task, subjects ate snacks with chopsticks and were allowed to drink water intermittently while using only the dominant hand. In the \textit{cleaning} task, subjects first sorted Lego pieces by color and then wiped the desk and surrounding objects using only the dominant hand.

\subsection{Data Preprocessing}
\label{subsec:data_preprocessing}
% bandpass filter, z-score normalization
% peak alignment, peak rejection, segmentation, stride
The CIME and Daily Activities datasets are used throughout the experiments for training and evaluation. For the Daily Activities dataset, peak alignment is performed before the main preprocessing pipeline. For each session, we estimate the average peak time offset between the band-pass filtered clean and noisy PPG recorded during the resting task. The offset is then used to temporally align each signal pair by shifting one signal relative to the other.

After peak alignment, both CIME and Daily Activities datasets are processed using the same preprocessing pipeline. Signals are segmented into overlapping 10-second windows with a stride of 0.25 s. A band-pass filter of 0.5--4.0 Hz is then applied. Next, peak rejection is performed on the clean windows to exclude low-quality segments, including those that may still contain residual motion artifacts despite being recorded from the relatively still hand. In the peak rejection step, peaks are detected from the clean PPG, and each window is retained only if both signal-quality criteria are satisfied, with the coefficient of variation of RR intervals limited to 0.1 or less and the coefficient of variation of peak amplitudes limited to 0.3 or less. Lastly, sample-wise z-score normalization is applied to each window.

\subsection{Experimental Settings}
\label{subsec:experimental_settings}

All experiments use the same optimization and model settings unless otherwise stated. For each experiment, the compared models are trained and evaluated separately on the Daily Activities dataset and the CIME dataset. MURFormer is trained with the AdamW optimizer using a learning rate of $4\times10^{-4}$ and a batch size of 256 for all three stages. Early stopping is applied with a patience of 20 epochs. The patch size of the patchifier is set to 10 for the Daily Activities dataset and 25 for the CIME dataset. The embedding dimension is 64, with 4 attention heads and a dropout rate of 0.1. The pass1 MACF backbone contains 8 layers, whereas the pass2 MACF backbone contains 4 layers. In the MACF block, the convolution kernel size is 5, the small-scale and large-scale partition sizes are 5 and 20, respectively.

Stages 1, 2, and 3 are trained using losses defined in Eqs.\eqref{eq:stage1_loss}, \eqref{eq:stage2_loss}, and \eqref{eq:stage3_loss}, respectively. For the Stage 1 and Stage 3 losses from Eq.\eqref{eq:stage1_loss} and Eq.\eqref{eq:stage3_loss}, $\alpha$, $\beta$, $\gamma$, and $\alpha_m$ are set to 5.  For Stage 1, the best model is selected based on the validation reconstruction error. For Stage 2, the best model is selected based on the validation AUROC for detecting high-error patches. For Stage 3, the best model is selected based on the validation loss. The uncertainty threshold $\tau$ used in Eq.\eqref{eq:core_mask} is determined from the validation set, where the threshold is set to the 95th percentile of validation uncertainty values.

For the Daily Activities dataset, we adopt a leave-one-subject-out (LOSO) setting. In each fold, one subject is reserved for testing, and the remaining subjects are used for training and validation. Specifically, Session 1 data from the remaining subjects are used for training and Session 2 data are used for validation. For the CIME dataset, we use data from 10 subjects for training and the remaining 38 subjects for testing following the original split\cite{xu2019photoplethysmography, xu2020stochastic}.

% These tables are positioned here to put this table in the same page as Results and Discussion Section
\begin{table*}[t]
\centering
\caption{Experiment 1: Comparison between the Transformer-based denoiser and the MACF-based denoiser (MURFormer pass1). Bold values indicate the better result within each dataset.}
\label{tab:exp1}
\resizebox{\textwidth}{!}{
\begin{tabular}{llccccccccc}
\toprule
Dataset & Model 
& SNR Mean 
& HR MAE 
& HR RMSE 
& PPI MAE 
& PPI RMSE 
& SDNN MAE 
& SDNN RMSE 
& RMSSD MAE 
& RMSSD RMSE \\
\midrule
Daily Activities & Transformer-based 
& 7.29 
& 1.88 
& 4.60 
& 15.93 
& 38.58 
& 35.76 
& 64.40 
& 54.97 
& 94.73 \\
& MACF-based 
& \textbf{8.45} 
& \textbf{1.52} 
& \textbf{4.17} 
& \textbf{12.34} 
& \textbf{32.67} 
& \textbf{14.28} 
& \textbf{30.21} 
& \textbf{18.54} 
& \textbf{36.51} \\
\midrule
CIME & Transformer-based 
& 10.81 
& 0.96 
& \textbf{3.14} 
& 9.36 
& \textbf{29.94} 
& 15.64 
& 34.35 
& 22.62 
& 46.69 \\
& MACF-based 
& \textbf{11.12} 
& \textbf{0.91} 
& 3.25 
& \textbf{8.79} 
& 30.20 
& \textbf{10.61} 
& \textbf{24.53} 
& \textbf{13.76} 
& \textbf{29.00} \\
\bottomrule
\end{tabular}
}
\end{table*}

\begin{table}[!t]
\centering
\caption{Parameter comparison between Transformer-based and MACF-based designs under the Daily Activities setting.}
\label{tab:exp1_params}
\begin{tabular}{lccc}
\toprule
Component & Transformer-based & MACF-based & Reduction (\%) \\
\midrule
Encoder block & 33,472 & 28,320 & 15.4\% \\
Full model & 273,601 & 232,321 & 15.1\% \\
\bottomrule
\end{tabular}
\end{table}

\subsection{Experiment 1: Transformer Encoder vs MACF Encoder}
\label{subsec:exp1}
% transformer vs MACF block
The first experiment compares the proposed MACF block with a standard Transformer encoder under the same reconstruction framework. Both models use the same patchifier--encoder--reconstruction head architecture as the pass1 reconstruction path of MURFormer, differing only in the encoder block, which is implemented as either a standard Transformer encoder or the proposed MACF block. For the MACF-based model, we use the Stage 1 model from Experiment 2 (Section~\ref{subsec:exp2}), which corresponds to the first-pass reconstruction path of MURFormer trained for direct reconstruction. The encoder depth is fixed to 8 layers in both models. This experiment evaluates whether MACF provides a more effective encoder for direct noisy-to-clean PPG reconstruction.

\subsection{Experiment 2: Direct Reconstruction vs Three-Stage Training}
\label{subsec:exp2}
% Stage 1 vs Stage 3
The second experiment is designed to justify the proposed three-stage training strategy and uncertainty-guided two-pass inference. We compare the Stage 1 model with the final model obtained after Stage 3 training. The Stage 1 model performs direct reconstruction using only the first-pass reconstruction path, whereas the final model applies two-pass inference after the full three-stage training. 

We also validate if the uncertainty head identifies unreliable regions accurately by treating it as a patch-level detector of high-error regions. On the held-out test set, patches whose Stage 1 reconstruction MSE falls in the top 10\% (\textit{q}=10\%) are labeled as positives, and AUROC/AUPRC are computed over all decision thresholds using predicted uncertainty as the score. For the Daily Activities dataset, AUROC and AUPRC are computed for each held-out subject and then averaged across all 24 folds.

\subsection{Experiment 3: Baseline Comparison}
\label{subsec:exp3}
% MURFormer vs Diffusion
% preprocessing differs from default exp setting
Finally, the third experiment compares MURFormer with the diffusion-based denoising model proposed by Xia et al.\cite{xia2025effective}, which was developed for the same task. The goal of this experiment is to directly compare the denoising performance of MURFormer against a recent baseline. Among recent learning-based PPG denoising methods, this model was selected as the primary baseline because it is reproducible and can be adapted to our experimental setting with clearly specified implementation details. Other recent methods \cite{sawangjai2025removal,liu2026flexima} are discussed as related work, but were not included as direct baselines due to limited reproducibility under the same training and evaluation protocol. In this experiment, max-absolute scaling which rescales the maximum absolute value to 1 is used instead of z-score normalization, following the original design of the diffusion baseline. This is because the diffusion model assumes bounded signal amplitudes during preprocessing and sampling, including clipping. Accordingly, to ensure a fair comparison, the input to MURFormer is also normalized using max-absolute scaling. 

In addition, to adapt the diffusion baseline to the datasets in the experiment, we adjusted the spectrogram configuration according to the sampling rate and window length while preserving a temporal resolution close to that of the original signal. For the Daily Activities dataset, we use a hop size of 10 samples, 40 mel frames, an FFT size of 64, 8 mel bands, and a maximum mel frequency of 5 Hz. For the CIME dataset, we use a hop size of 25 samples and an FFT size of 128, while keeping the remaining settings identical to those used for the Daily Activities dataset. All other preprocessing steps follow the settings stated in Section~\ref{subsec:data_preprocessing}. The training settings of the diffusion baseline follow those of the original work, with a batch size of 128, a learning rate of 0.001, and 100{,}000 training iterations. MURFormer is trained using the same settings described in Section~\ref{subsec:experimental_settings}.

% Tables for Results Section.
% <Table> Stage 1 vs Stage 3
\begin{table*}[t]
\centering
\caption{Experiment 2: Comparison among the noisy input, Stage 1 direct reconstruction, and the final Stage 3 model of MURFormer. Bold values indicate the better result within each dataset.}
\label{tab:exp2}
\resizebox{\textwidth}{!}{
\begin{tabular}{llccccccccc}
\toprule
Dataset & Model
& SNR Mean
& HR MAE
& HR RMSE
& PPI MAE
& PPI RMSE
& SDNN MAE
& SDNN RMSE
& RMSSD MAE
& RMSSD RMSE \\
\midrule
& Noisy
& 3.72
& 4.02
& 7.47
& 34.49
& 65.95
& 79.77
& 115.50
& 116.85
& 166.66 \\
Daily Activities & Stage 1
& 8.45
& 1.52
& 4.17
& 12.34
& 32.67
& 14.28
& 30.21
& 18.54
& 36.51 \\
& Stage 3
& \textbf{8.51}
& \textbf{1.38}
& \textbf{3.97}
& \textbf{11.31}
& \textbf{31.50}
& \textbf{10.94}
& \textbf{23.41}
& \textbf{13.86}
& \textbf{26.27} \\
\midrule
& Noisy
& 5.44
& 4.53
& 9.07
& 41.98
& 81.71
& 60.65
& 90.54
& 87.52
& 127.70 \\
CIME & Stage 1
& 11.12
& 0.91
& 3.25
& 8.79
& \textbf{30.20}
& 10.61
& 24.53
& 13.76
& 29.00 \\
& Stage 3
& \textbf{11.21}
& \textbf{0.88}
& \textbf{3.23}
& \textbf{8.58}
& 30.28
& \textbf{9.52}
& \textbf{22.36}
& \textbf{11.99}
& \textbf{25.00} \\
\bottomrule
\end{tabular}
}
\end{table*}

\section{Results and Discussion}
\label{sec:results}

In this section, we demonstrate that the proposed MACF block and uncertainty-guided two-pass framework enable MURFormer to effectively denoise motion-corrupted signals and improve denoising performance across the evaluated settings. We evaluate and compare denoising performance using signal-to-noise ratio (SNR) and physiological metrics derived from the reconstructed PPG. In addition to heart rate (HR), we report peak-to-peak interval (PPI), standard deviation of RR intervals (SDNN), and root mean square of successive differences (RMSSD) to assess whether the denoised PPG preserves beat-to-beat timing and heart-rate-variability (HRV) measures beyond waveform-level signal quality. These metrics are extracted from the denoised PPG and compared with the corresponding values obtained from the paired clean reference PPG. For HR, PPI, SDNN, and RMSSD, both mean absolute error (MAE) and root mean square error (RMSE) are reported. HR-related errors are expressed in beats per minute, whereas PPI, SDNN, and RMSSD-related errors are expressed in milliseconds.

% Exp1
\subsection{Experiment 1 Results: Transformer Encoder vs MACF Encoder}
% Transformer Denoiser vs MURFormer pass1, SNR/HR MAE better
Table~\ref{tab:exp1} compares the direct reconstruction performance of the Transformer-based denoiser and the MACF-based denoiser, where the latter corresponds to the pass1 reconstruction path of MURFormer. On the Daily Activities dataset, the MACF-based model outperformed the Transformer-based model across all reported metrics, with statistically significant improvements in both SNR and HR MAE across 24 subjects according to the Wilcoxon signed-rank test(SNR, $p = 1.19 \times 10^{-7}$; HR, $p = 4.30 \times 10^{-4}$), as well as substantially lower PPI and HRV-related errors. Similar improvements were observed on the CIME dataset, where the MACF-based model increased SNR from 10.81 to 11.12 dB, reduced HR MAE from 0.96 to 0.91 bpm, and showed consistently lowered PPI and HRV-related errors. This indicates that the proposed MACF block provides a more effective encoder for direct noisy-to-clean reconstruction. Notably, this gain was achieved with approximately 15\% fewer parameters than the Transformer-based denoiser as shown in Table~\ref{tab:exp1_params}. This reduction in overall model size is mainly attributed to the MACF block, which is more parameter-efficient than a standard Transformer encoder under the same embedding dimension. These results support the effectiveness of MACF as a parameter-efficient encoder for noisy-to-clean PPG reconstruction.
% (check full version in the old file)

% Exp2
\subsection{Experiment 2 Results: Direct Reconstruction vs Three-Stage Training}
\label{subsec:exp2_results}

\begin{figure}[t]
    \centering
    \includegraphics[width=0.8\linewidth]{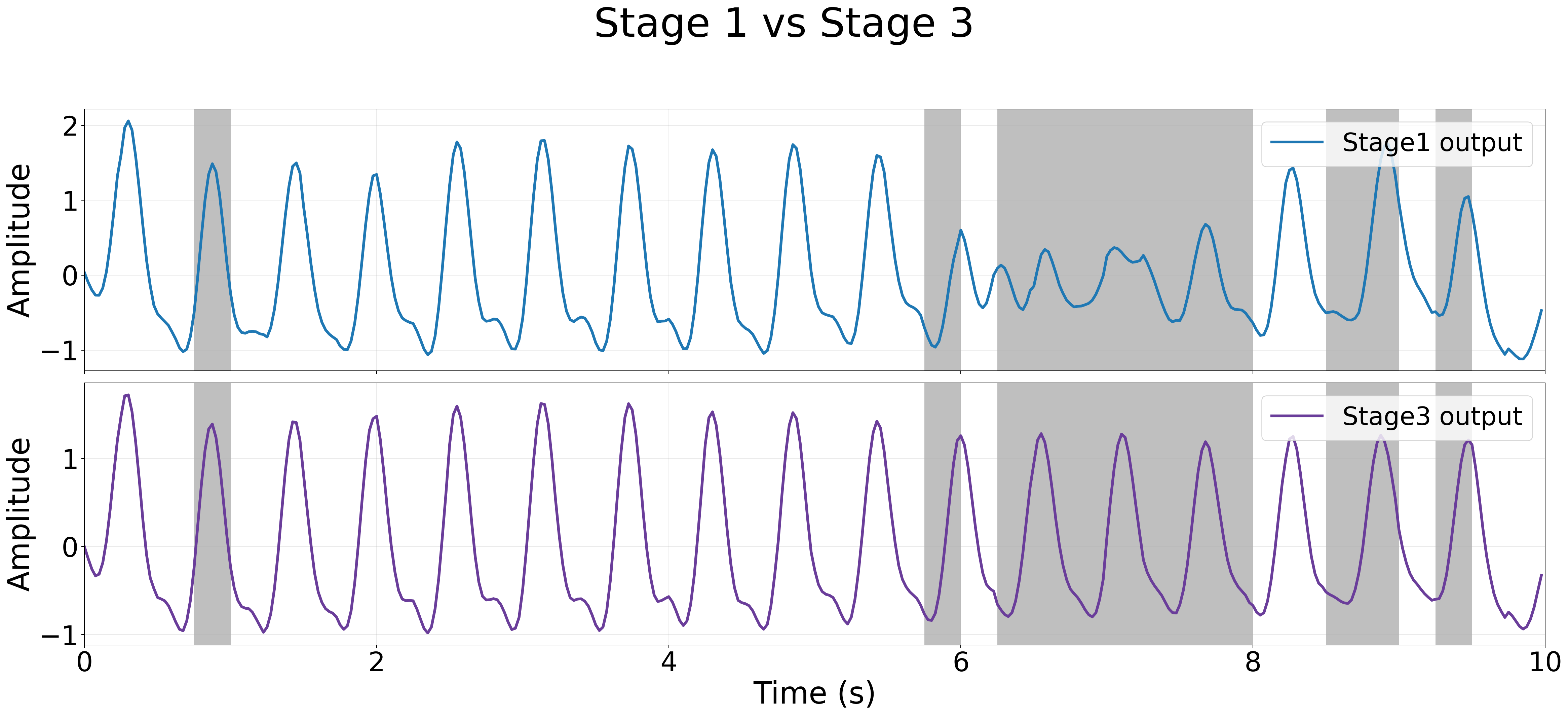}
    \caption{Comparison of Stage 1 and Stage 3 outputs. Gray regions indicate patches exceeding the uncertainty threshold, highlighting unreliable regions of Stage 1 identified by the uncertainty head.}
    \label{fig:stage1vs3}
\end{figure}

\begin{table}[t]
\centering
\caption{Uncertainty validation results on test set. High-error patch detection AUROC/AUPRC (\textit{q}=10\%)}
\label{tab:uncertainty}
\begin{tabular}{lcc}
\toprule
Dataset & AUROC(\textit{q}=10\%) & AUPRC(\textit{q}=10\%) \\
\midrule
Daily Activities & 0.79 & 0.33 \\
CIME & 0.79 & 0.36 \\
\bottomrule
\end{tabular}
\end{table}

Table~\ref{tab:exp2} presents the comparison between the Stage 1 direct-reconstruction model with the final Stage 3 model of MURFormer, which applies the full uncertainty-guided two-pass framework after three-stage training.

% revised
The final Stage 3 model consistently improved upon the Stage 1 direct-reconstruction model on both datasets. Although the absolute improvement was modest, the improvements in SNR and HR MAE on the Daily Activities dataset were statistically significant according to the Wilcoxon signed-rank test (SNR, $p = 0.029$; HR, $p = 6.50 \times 10^{-4}$). The Stage 3 model outperformed the Stage 1 model across all reported metrics on both datasets except for a slight increase in PPI RMSE on CIME, supporting the role of the uncertainty-guided second-pass refinement.

% Original paragraph split into methods and Results (Experiment2)
To verify the usefulness of the uncertainty-guided two-pass approach in denoising, Table~\ref{tab:uncertainty} reports the high-error patch detection performance on the test set. The uncertainty head achieved an AUROC of 0.79 on both datasets, with AUPRC values of 0.33 and 0.36 on the Daily Activities and CIME datasets, respectively. Since the positive class ratio is 10\%, these AUPRC values are substantially above the random baseline, indicating that the predicted uncertainty is reasonably aligned with the actual reconstruction error after the first pass, which supports its use in Stage 3 refinement.

Fig.~\ref{fig:stage1vs3} shows a representative example of how the proposed method refines the initial denoised output for a single noisy sample. In the figure, the blue curve denotes the Stage 1 output and the purple curve denotes the Stage 3 output. Gray regions correspond to patches whose uncertainty values exceed the uncertainty threshold according to the uncertainty map predicted by the uncertainty head. Although the Stage 1 reconstruction output captures the overall waveform, some local regions remain imperfectly restored. The uncertainty head identifies these regions and the Stage 3 output further corrects them.

% Shortened from the original paragraph
These results suggest that the first-pass MACF reconstruction removes most artifacts, while the uncertainty-guided second pass provides additional localized correction with the help of the uncertainty head.

\begin{figure}[!t]
    \centering

    \begin{subfigure}{0.8\linewidth}
        \centering
        \includegraphics[width=\linewidth]{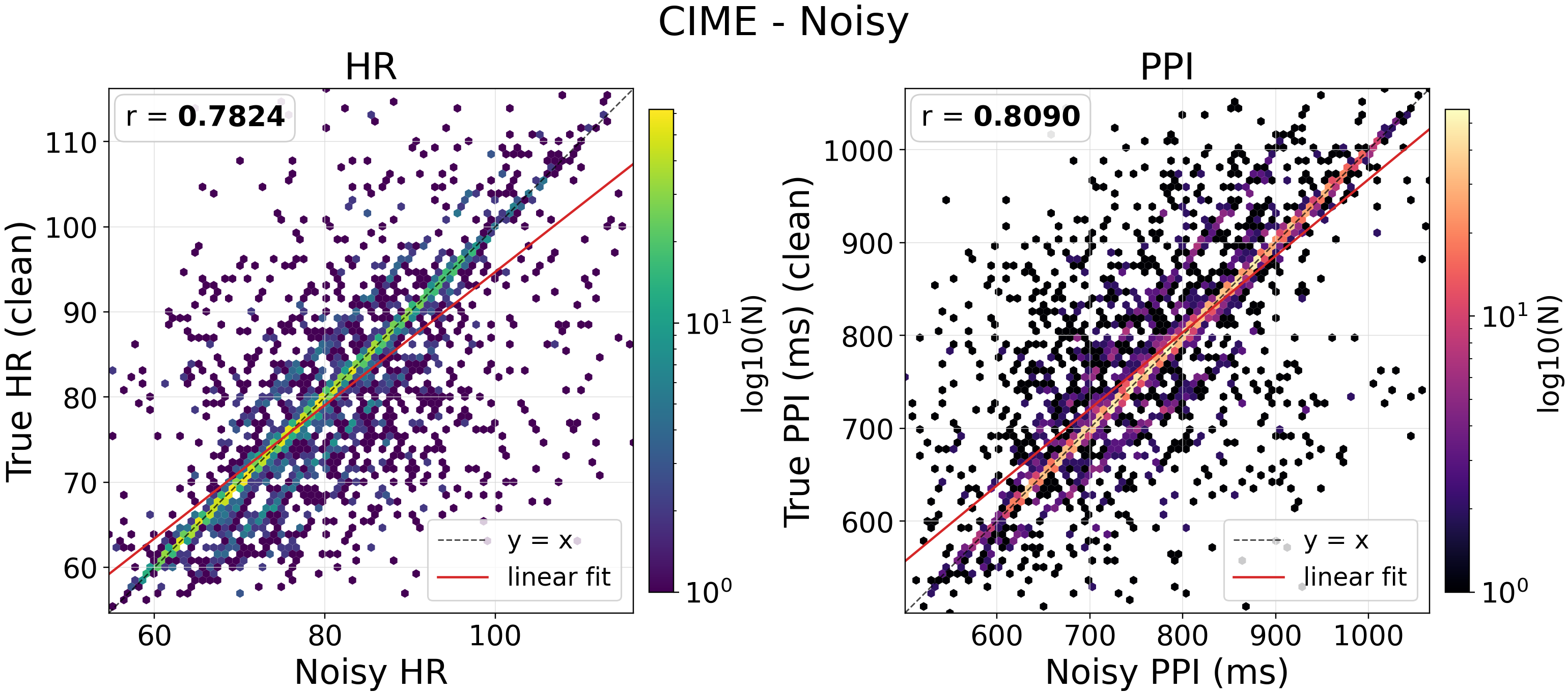}
        \caption{}
        \label{fig:cime_noisy}
    \end{subfigure}

    \vspace{0.5em}

    \begin{subfigure}{0.8\linewidth}
        \centering
        \includegraphics[width=\linewidth]{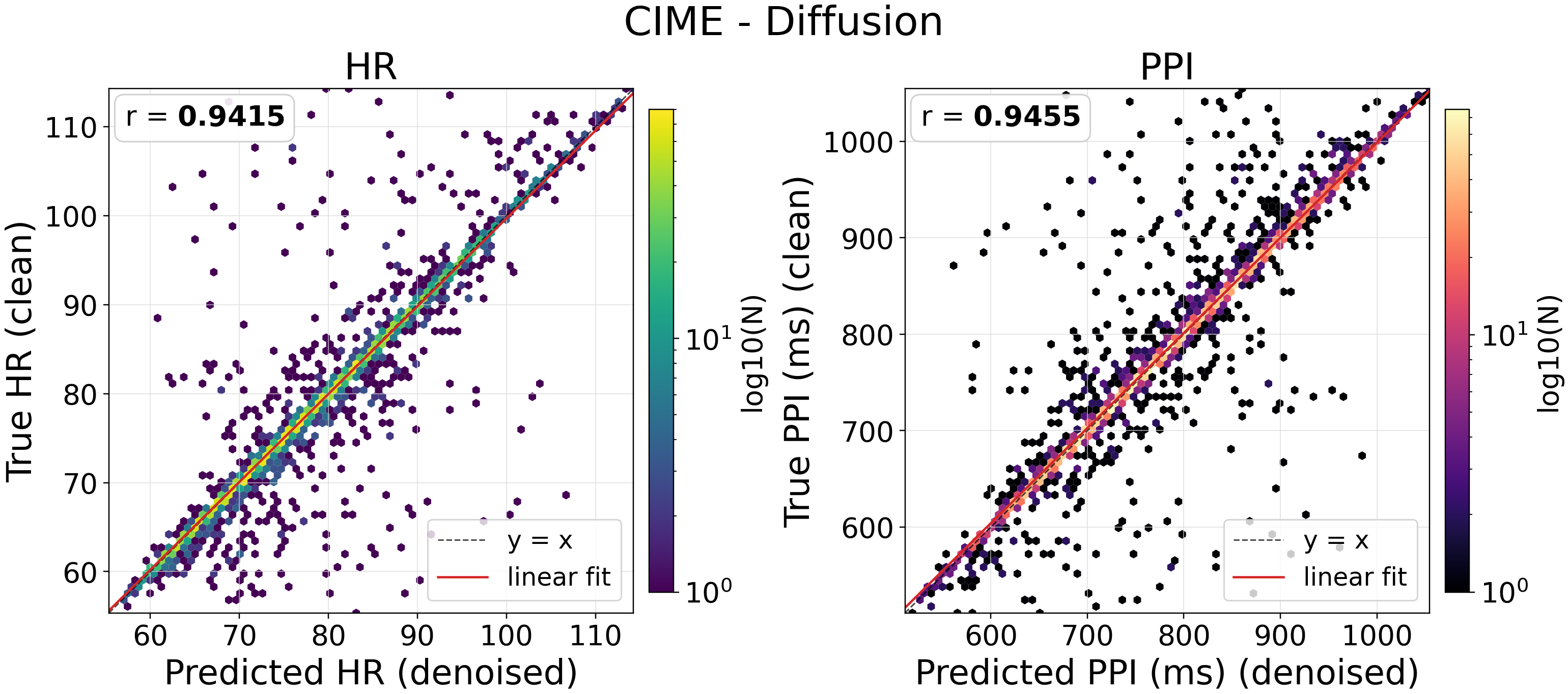}
        \caption{}
        \label{fig:cime_diffusion}
    \end{subfigure}

    \vspace{0.5em}

    \begin{subfigure}{0.8\linewidth}
        \centering
        \includegraphics[width=\linewidth]{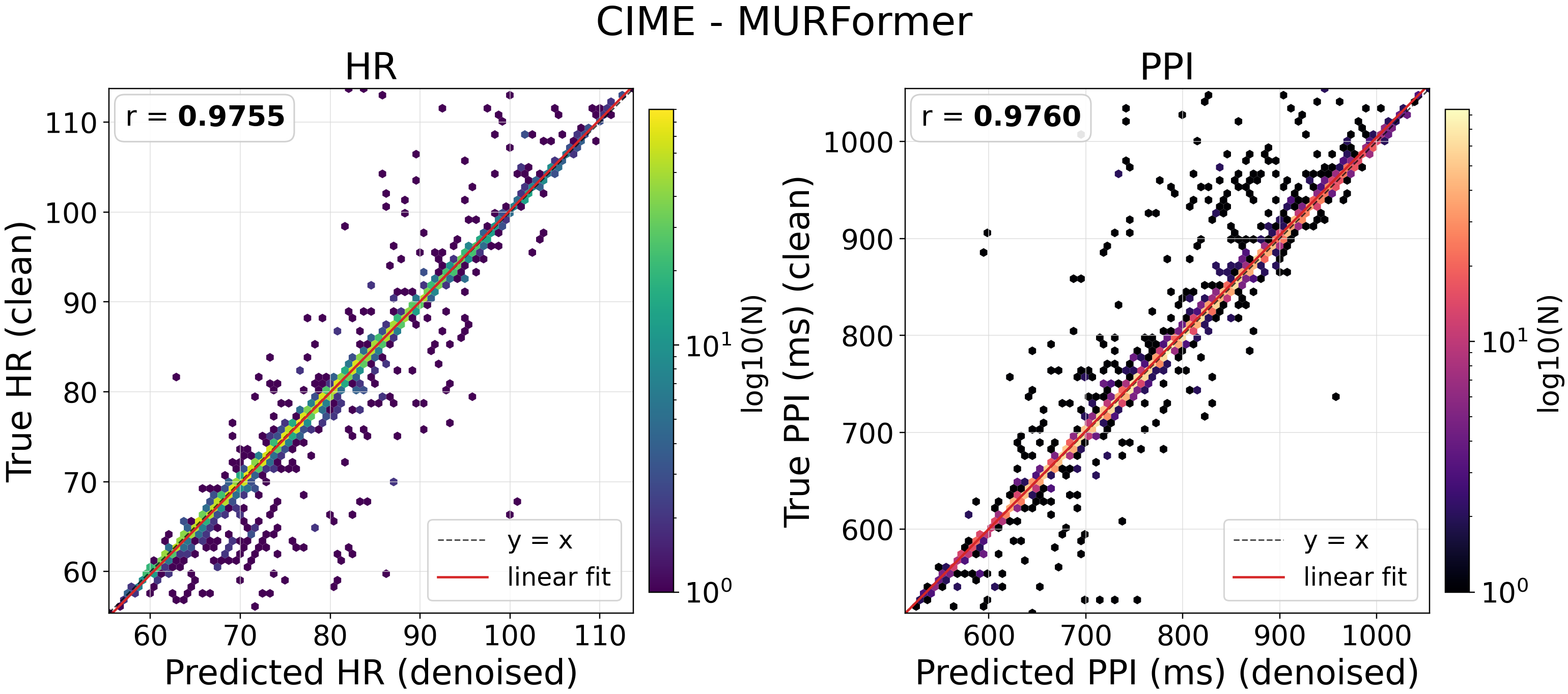}
        \caption{}
        \label{fig:cime_murformer}
    \end{subfigure}

    \caption{Pearson correlation plots of HR and PPI estimates on the CIME dataset. The plots compare estimates derived from (a) noisy PPG, (b) diffusion-denoised PPG, and (c) MURFormer-denoised PPG against the clean PPG-derived reference values.}
    \label{fig:cime_comparison}
\end{figure}

\begin{table*}[t]
\centering
\caption{Experiment 3: Comparison among the noisy input, the diffusion-based baseline, and MURFormer with max-absolute scaling. Bold values indicate the best result within each dataset.}
\label{tab:exp3}
\resizebox{\textwidth}{!}{
\begin{tabular}{llccccccccc}
\toprule
Dataset & Model
& SNR Mean
& HR MAE
& HR RMSE
& PPI MAE
& PPI RMSE
& SDNN MAE
& SDNN RMSE
& RMSSD MAE
& RMSSD RMSE \\
\midrule
Daily Activities & Noisy
& 4.04
& 4.02
& 7.47
& 34.49
& 65.95
& 79.77
& 115.50
& 116.85
& 166.66 \\
& Diffusion
& 7.47
& 1.64
& 4.77
& 13.67
& 39.56
& \textbf{10.61}
& \textbf{22.35}
& 14.23
& 28.41 \\
& MURFormer
& \textbf{8.16}
& \textbf{1.37}
& \textbf{3.92}
& \textbf{11.20}
& \textbf{31.09}
& 11.07
& 23.66
& \textbf{14.04}
& \textbf{26.89} \\
\midrule
CIME & Noisy
& 5.53
& 4.53
& 9.07
& 41.99
& 81.71
& 60.65
& 90.54
& 87.52
& 127.70 \\
& Diffusion
& 9.60
& 1.41
& 4.47
& 13.58
& 42.10
& 11.76
& 24.90
& 14.42
& 27.63 \\
& MURFormer
& \textbf{10.72}
& \textbf{0.82}
& \textbf{2.96}
& \textbf{7.99}
& \textbf{28.03}
& \textbf{9.16}
& \textbf{21.04}
& \textbf{11.42}
& \textbf{22.83} \\
\bottomrule
\end{tabular}
}
\end{table*}

% Exp3
\subsection{Experiment 3 Results: Baseline Comparison}
% Diffusion vs MURFormer, 

% shortened this Section. (check full version in the old file)
Table~\ref{tab:exp3} compares the denoising performance of MURFormer and the diffusion-based baseline\cite{xia2025effective} under the same max-absolute scaling setting. Both models improved substantially over the noisy input, but MURFormer achieved better overall performance on both datasets. On the Daily Activities dataset, MURFormer yielded higher SNR, lower HR, PPI, and RMSSD errors than the diffusion baseline. On the CIME dataset, MURFormer outperformed the diffusion baseline across all reported metrics, significantly increasing SNR from 9.60 to 10.72 dB and reducing HR MAE from 1.41 to 0.82 bpm (SNR, $p = 3.66 \times 10^{-5}$; HR, $p = 0.042$). It also achieved lower errors for PPI, SDNN, and RMSSD, indicating improved preservation of both beat-to-beat timing and HRV-related characteristics.

Fig.~\ref{fig:cime_comparison} further supports this result by showing Pearson correlation plots for HR and PPI on the CIME dataset. Compared with the diffusion-denoised output, the MURFormer-denoised output showed stronger agreement with the clean PPG-derived HR and PPI reference values. MURFormer achieved higher correlations for both HR and PPI (Fig.~\ref{fig:cime_murformer}, $r=0.9755$ and $r=0.9760$) than the diffusion baseline (Fig.~\ref{fig:cime_diffusion}, $r=0.9416$ and $r=0.9460$). The MURFormer estimates were more tightly concentrated around the identity line.

\subsection{Comparison with Noisy Input}

We next examine the improvement of MURFormer over the original noisy input using the standard preprocessing setting in Table~\ref{tab:exp2}. On the Daily Activities dataset, MURFormer (Stage 3) increased SNR from 3.72 to 8.51 dB and reduced HR MAE from 4.02 to 1.38 bpm. On the CIME dataset, SNR increased from 5.44 to 11.21 dB while HR MAE decreased from 4.53 to 0.88 bpm. These improvements were accompanied by consistent reductions in PPI, SDNN, and RMSSD errors, indicating that MURFormer substantially improves waveform-level signal quality and beat timing and HRV-related physiological fidelity.

The correlation plots in Fig.~\ref{fig:cime_comparison} also show that MURFormer-denoised estimates (Fig.~\ref{fig:cime_murformer}) align more closely with the clean reference than estimates derived from the noisy PPG (Fig.~\ref{fig:cime_noisy}).

\begin{figure}[t]
    \centering
    \includegraphics[width=0.8\linewidth]{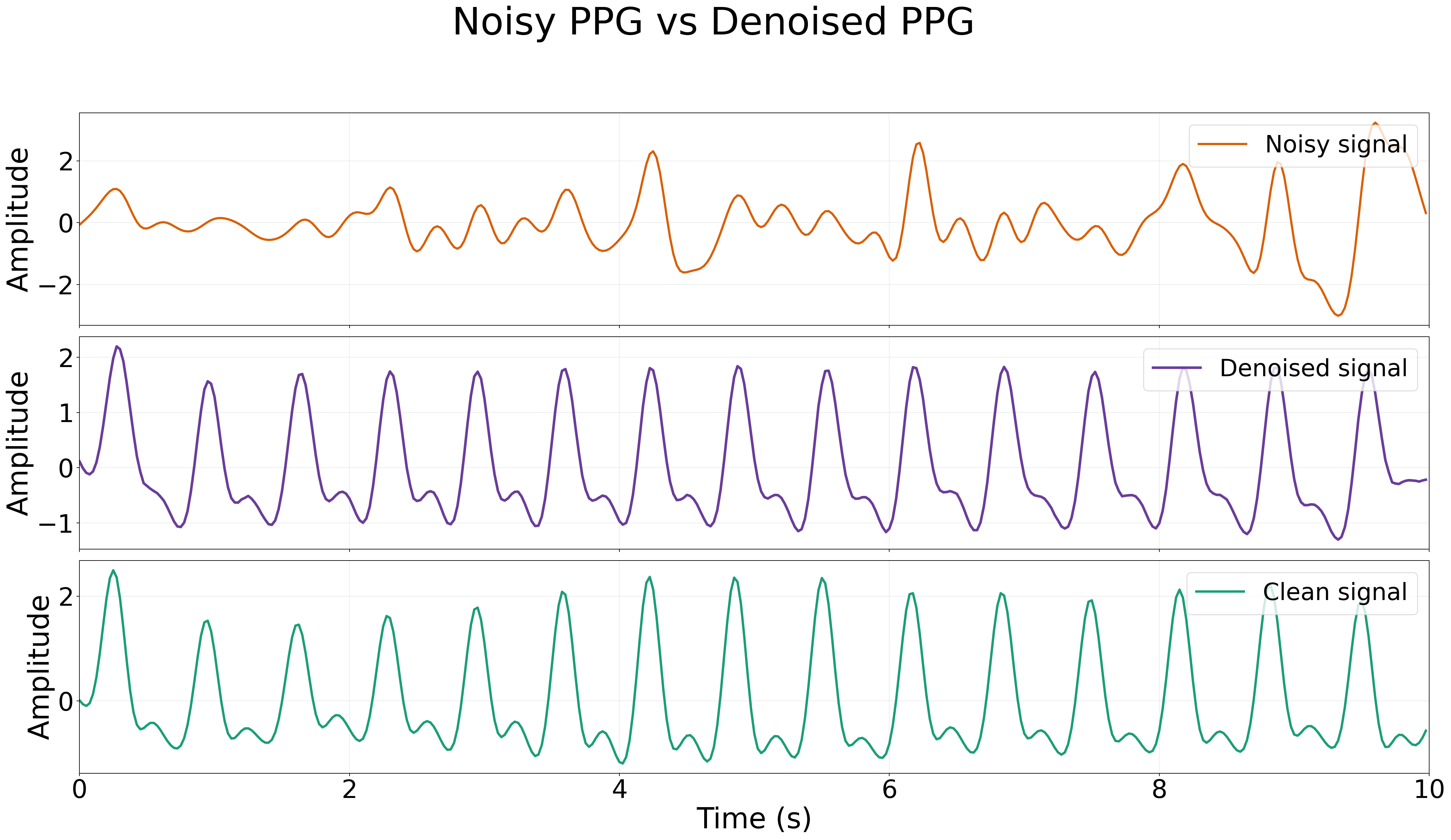}
    \caption{Comparison of noisy PPG, denoised output, and clean PPG.}
    \label{fig:noisyvsdenoised}
\end{figure}

A qualitative example shown in Fig.~\ref{fig:noisyvsdenoised} illustrates how MURFormer recovers the PPG waveform even when the input is heavily corrupted. Notably, the model recovers not only the major systolic peaks but also the smaller diastolic peaks, suggesting that the proposed framework can preserve the fine-grained morphological structures of the PPG waveform.

\begin{table}[t]
\centering
\caption{Comparison of the number of model parameters among recent PPG denoising models and MURFormer.}
\label{tab:model_parameters}
\begin{tabular}{lc}
\toprule
\textbf{Model} & \textbf{Number of Parameters} \\
\midrule
FlexiMA \cite{liu2026flexima} & 109.8M \\
CycleGAN \cite{afandizadeh2023accurate} & 94.35M \\
Diffusion \cite{xia2025effective} & 2.3M \\
MURFormer & 0.36M \\
\bottomrule
\end{tabular}
\end{table}

\subsection{MURFormer on Varying Input Window Sizes}
\label{subsec:window_size_results}

MURFormer is designed to accept variable-length input sequences. We further evaluate the robustness of MURFormer to varying input lengths by using input windows of different durations at test time. The models trained using 10-s windows from a previous experiment in Section~\ref{subsec:exp2_results} are directly applied to test inputs of 5 s, 10 s, 30 s, and 60 s. For the Daily Activities dataset, models trained under the leave-one-subject-out protocol were reused, and each model was evaluated on its corresponding held-out subject using different input window sizes. For CIME, the 60-s setting was not considered because each trial is 30 s long.
% 5s (1s stride), 30s (10s), 60s(10s)

\begin{figure}[!t]
    \centering
    \includegraphics[width=\linewidth]{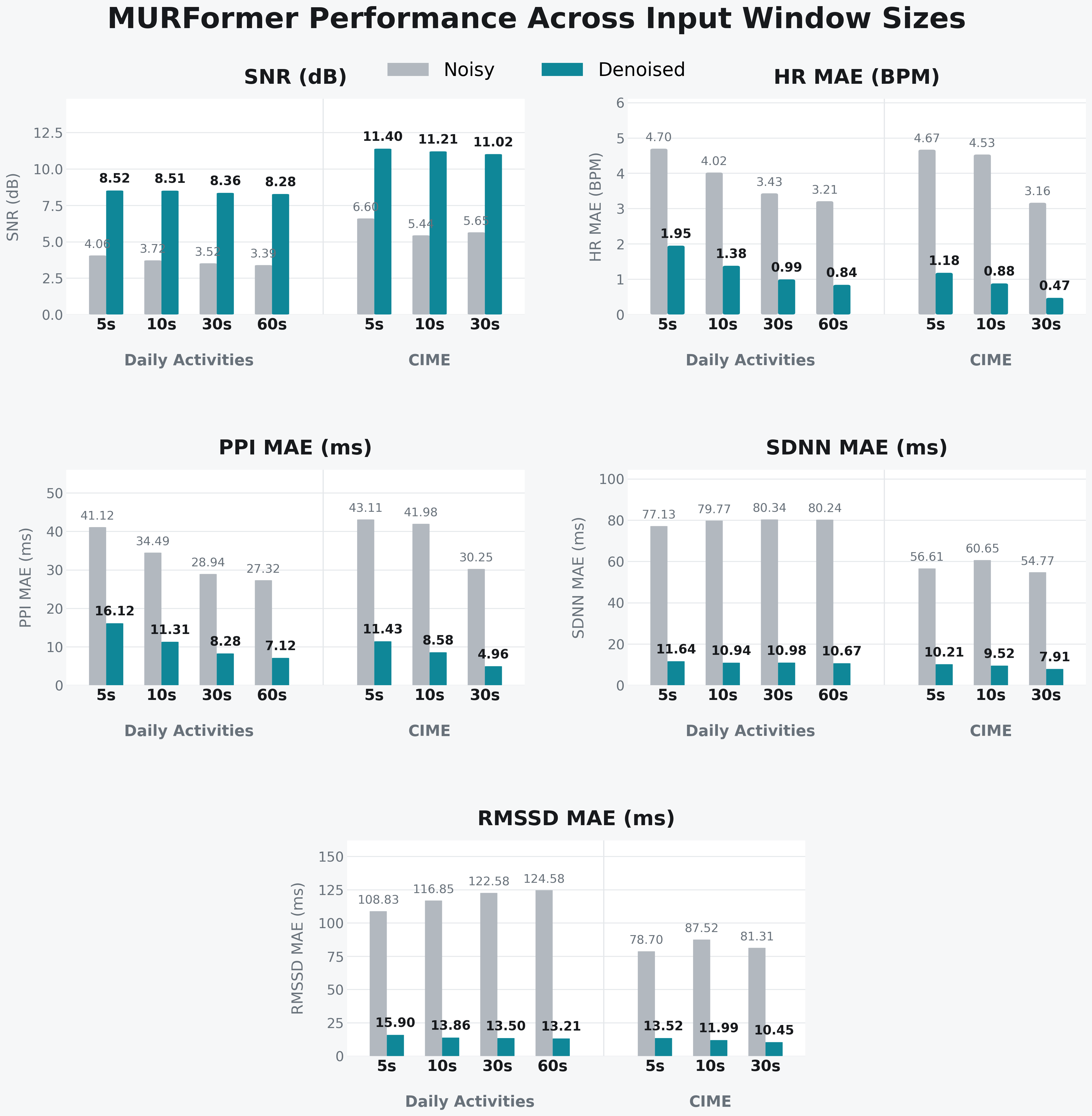}
    \caption{Comparison of noisy PPG and denoised output of MURFormer across different input window sizes at test time.}
    \label{fig:murformer_windows}
\end{figure}

Fig.~\ref{fig:murformer_windows} illustrates the SNR, HR, PPI, SDNN, and RMSSD MAEs of the noisy PPG signals and the MURFormer-denoised outputs on each dataset across different input window sizes. The results show that MURFormer maintains strong denoising performance across all tested input lengths. For example, on the Daily Activities dataset, HR MAE decreased from 4.70 to 1.95 bpm at 5 s and from 3.21 to 0.84 bpm at 60 s. On CIME, HR MAE decreased from 4.67 to 1.18 bpm at 5 s and from 3.16 to 0.47 bpm at 30 s. Consistent trends were observed across the remaining metrics, as MURFormer improved SNR and reduced PPI, SDNN, and RMSSD errors in all input lengths tested.

These results indicate that MURFormer generalizes well to a range of input lengths at inference time even when it was trained using 10-s windows. While the best SNR is obtained near the original training scale, longer input windows tend to provide lower HR and HRV-related errors, likely because they provide more temporal context for stable physiological estimation. Overall, the proposed method shows robust denoising performance under varying input window sizes, suggesting that the learned representation is not tightly restricted to a single input duration.

\subsection{Model Size}
\label{subsec:model_size}
Table~\ref{tab:model_parameters} compares the number of parameters of MURFormer with those of recent learning-based PPG denoising models. For the diffusion-based model~\cite{xia2025effective}, the parameter count was estimated from the open-source implementation used in our baseline experiment. For CycleGAN~\cite{afandizadeh2023accurate} and FlexiMA~\cite{liu2026flexima}, whose implementations are not publicly available, we report the parameter counts stated in the paper. MURFormer uses approximately 6.4 times fewer parameters than the diffusion baseline and has a  substantially smaller parameter count than FlexiMA and CycleGAN. This result highlights the compactness of MURFormer relative to recent learning-based PPG denoising models.

\section{Conclusion}
\label{sec:conclusion}
% main points, weakness 
This study proposes MURFormer, a lightweight uncertainty-guided two-pass framework for PPG denoising. Experimental results on the self-collected Daily Activities dataset under a cross-subject evaluation protocol and on the CIME dataset demonstrate that MURFormer improves signal quality and physiological fidelity while outperforming the diffusion-based baseline. The model also maintained strong denoising performance across varying input window lengths. These gains were achieved with a substantially smaller number of parameters than recent models, supporting its suitability for practical deployment. Moreover, the proposed MACF block was designed to reflect the temporal and multi-scale dependencies of PPG, and its favorable balance between parameter efficiency and reconstruction performance in PPG denoising suggests that it may serve as a useful basic building block for broader PPG-related tasks beyond denoising.

\section*{Conflict of Interest Statement}
The authors declare that they have no known financial or personal conflicts of interest that could have influenced the work reported in this paper.

% \appendices

\section*{References}
\bibliographystyle{IEEEtran}
\bibliography{references}

\end{document}